\documentclass[11pt,letterpaper]{article}
\usepackage{emnlp2016}
\usepackage{times}
\usepackage{latexsym}

\usepackage{url}
\usepackage{graphicx} % more modern
\usepackage{subfigure} 
\usepackage[flushleft]{threeparttable}
\usepackage{color}

\usepackage{algorithm}
\usepackage{algorithmic}

\usepackage{etoolbox}
\newcommand{\algorithmicdoinparallel}{\textbf{do in parallel}}
\makeatletter
\AtBeginEnvironment{algorithmic}{%
  \newcommand{\FORALLP}[2][default]{\textbf{for all machine} #2\ %
    \algorithmicdoinparallel\ALC@com{#1}\begin{ALC@for}}%
}

\emnlpfinalcopy

\def\emnlppaperid{***}

\input{Definitions}

\title{\wrr: Learning Word Embeddings via Robust Ranking}

\author{Shihao Ji \\
Parallel Computing Lab, Intel\\
\texttt{shihao.ji@intel.com} \\
\And
Hyokun Yun \\
Amazon \\
\texttt{yunhyoku@amazon.com} \\
\And
Pinar Yanardag \\
Purdue University \\
\texttt{ypinar@purdue.edu} \\
\AND
Shin Matsushima \\
University of Tokyo \\
\texttt{shin\_matsushima@mist.} \\
\texttt{i.u-tokyo.ac.jp} \\
\And
S. V. N. Vishwanathan \\
Univ. of California, Santa Cruz \\
\texttt{vishy@ucsc.edu}}

\date{}

\begin{document}

\maketitle
\begin{abstract}
  Embedding words in a vector space has gained a lot of attention in 
  recent years. While state-of-the-art methods provide
  efficient computation of word similarities via a low-dimensional
  matrix embedding, their motivation is often left unclear. In this
  paper, we argue that word embedding can be naturally viewed as a
  \textit{ranking} problem due to the ranking nature of the evaluation
  metrics. Then, based on this insight, we propose a
  novel framework \wrr that efficiently estimates word representations
  via robust ranking, in which the attention mechanism and robustness 
  to noise are readily achieved via the DCG-like ranking losses.
  The performance of \wrr is measured in word similarity and word 
  analogy benchmarks, and the results are compared to the state-of-the-art 
  word embedding techniques. Our algorithm is very competitive to the state-of-the-
  arts on large corpora, while outperforms them by a significant margin when the
  training set is limited (\emph{i.e.}, sparse and noisy). With 17 million tokens, 
  \wrr performs almost as well as existing methods using 7.2 billion tokens on a 
  popular word similarity benchmark. Our multi-node distributed implementation of 
  \wrr is publicly available for general usage.
\end{abstract}

\section{Introduction}
\label{sec:Introduction}

Embedding words into a vector space, such that \emph{semantic} and
\emph{syntactic} regularities between words are preserved, is an
important sub-task for many applications of natural language
processing. \newcite{MikCheCorDea13} generated considerable excitement in
the machine learning and natural language processing communities by
introducing a neural network based model, which they call \wv. It was
shown that \wv produces state-of-the-art performance on both word
similarity as well as word analogy tasks. The word similarity task is to
retrieve words that are similar to a given word. On the other hand, word
analogy requires answering queries of the form \emph{a:b;c:?}, where
\emph{a, b}, and \emph{c} are words from the vocabulary, and the answer
to the query must be semantically related to \emph{c} in the same way as
\emph{b} is related to \emph{a}. This is best illustrated with a
concrete example: Given the query \emph{king:queen;man:?} we expect the
model to output \emph{woman}.

The impressive performance of \wv led to a flurry of papers, which tried
to explain and improve the performance of \wv both theoretically
\cite{AroLiLiaMaetal15} and empirically \cite{LevGol14}. One
interpretation of \wv is that it is approximately maximizing the
positive pointwise mutual information (PMI), and \newcite{LevGol14}
showed that directly optimizing this gives good results. On the other
hand, \newcite{PenSocMan14} showed performance comparable to \wv by using
a modified matrix factorization model, which optimizes a log loss.

Somewhat surprisingly, \newcite{LevGolDag15} showed that much of the
performance gains of these new word embedding methods are due to certain
hyperparameter optimizations and system-design choices. In other words,
if one sets up careful experiments, then existing word embedding models
more or less perform comparably to each other. We conjecture that this
is because, at a high level, all these methods are based on the
following template: From a large text corpus eliminate infrequent words,
and compute a $\abr{\Wcal} \times \abr{\Ccal}$ word-context
co-occurrence count matrix; a context is a word which appears less than
$d$ distance away from a given word in the text, where $d$ is a tunable
parameter. Let $w \in \Wcal$ be a word and $c \in \Ccal$ be a context,
and let $X_{w, c}$ be the (potentially normalized) co-occurrence
count. One learns a function $f(w,c)$ which approximates a transformed
version of $X_{w,c}$. Different methods differ essentially in the
transformation function they use and the parametric form of $f$
\cite{LevGolDag15}. For example, \gl \cite{PenSocMan14} uses
$f\rbr{w, c} = \inner{\ub_{w}}{\vb_{c}}$ where $\ub_{w}$ and $\vb_{c}$
are $k$ dimensional vectors, $\inner{\cdot}{\cdot}$ denotes the
Euclidean dot product, and one approximates
$f\rbr{w, c} \approx \log X_{w, c}$. On the other hand, as
\newcite{LevGol14} show, \wv can be seen as using the same $f(w, c)$ as
\gl but trying to approximate
$f\rbr{w, c} \approx \text{PMI}(X_{w, c}) - \log n$, where
$\text{PMI}(\cdot)$ is the pairwise mutual information \cite{CovTho91} and $n$ is the number of negative samples.

In this paper, we approach the word embedding task from a different
perspective by formulating it as a \emph{ranking} problem. That is,
given a word $w$, we aim to output an ordered list
$\rbr{c_1, c_2, \cdots }$ of context words from $\Ccal$ such that words
that co-occur with $w$ appear at the top of the list.  If $\rank(w, c)$
denotes the rank of $c$ in the list, then typical ranking losses
optimize the following objective:
$\sum_{(w,c) \in \Omega} \rho\rbr{\rank(w, c)}$, where
$\Omega \subset \Wcal \times \Ccal$ is the set of word-context pairs
that co-occur in the corpus, and $\rho(\cdot)$ is a ranking loss function that is monotonically increasing and concave (see Sec.~\ref{sec:WordEmbeddingvia} for a justification).

Casting word embedding as ranking has two distinctive
advantages.  First, our method is \emph{discriminative} rather than
generative; in other words, instead of modeling (a transformation of)
$X_{w,c}$ directly, we only aim to model the relative order of
$X_{w, \cdot}$ values in each row. This formulation fits naturally to
popular word embedding tasks such as word similarity/analogy since instead of
the likelihood of each word, we are interested in finding the most relevant words in a given
context\footnote{Roughly speaking, this difference in viewpoint is analogous to the difference between pointwise loss function vs listwise loss function used in ranking \cite{LeeLin13}.}. Second, casting word embedding as a ranking problem enables us
to design models robust to noise \cite{YunRamVis14} and focusing more on differentiating top
relevant words, a kind of attention mechanism that has been proved very 
useful in deep learning \cite{LarHin10,MniHeeGra14,BahChoBen15}. Both issues are very critical in the domain of word embedding since (1) the
co-occurrence matrix might be noisy due to grammatical errors or
unconventional use of language, \emph{i.e.}, certain words might
co-occur purely by chance, a phenomenon more acute in
smaller document corpora collected from diverse sources; and (2) it's very challenging to sort out a few most relevant words from a very large vocabulary, thus some kind of attention mechanism that can trade off the resolution on most relevant words with the resolution on less relevant words is needed. We will show in the experiments that our method can mitigate some of these issues; with 17 million tokens our method performs almost as well as existing
methods using 7.2 billion tokens on a popular word similarity benchmark.

% The rest of the paper is organized as follows: In Section
% \ref{sec:WordEmbeddingvia} we derive a general ranking loss, and show
% how it can be optimized efficiently. Related work is discussed in
% Section~\ref{sec:RelatedWork}. Empirical results are presented in
% section \ref{sec:Experiments}, and the paper concludes with a discussion
% in section \ref{sec:Discussion}.

\section{Word Embedding via Ranking}
\label{sec:WordEmbeddingvia}

\subsection{Notation}
\label{sec:Notation}

We use $w$ to denote a word and $c$ to denote a context. The set of
all words, that is, the vocabulary is denoted as $\Wcal$ and the set
of all context words is denoted $\Ccal$. We will use $\Omega \subset
\Wcal \times \Ccal$ to denote the set of all word-context pairs that
were observed in the data, $\Omega_{w}$ to denote the set of contexts
that co-occured with a given word $w$, and similarly $\Omega_{c}$ to denote the
words that co-occurred with a given context $c$. The size of a set is
denoted as $\abr{\cdot}$. The inner product between vectors is denoted
as $\inner{\cdot}{\cdot}$.

\subsection{Ranking Model}
\label{sec:RankingModel}

Let $\ub_w$ denote the $k$-dimensional embedding of a word $w$, and
$\vb_c$ denote that of a context $c$. For convenience, we collect
embedding parameters for words and contexts as $\Ub := \cbr{\ub_w}_{w
  \in \Wcal}$, and $\Vb := \cbr{\vb_c}_{c \in \Ccal}$.

We aim to capture the relevance of context $c$ for word $w$ by the
inner product between their embedding vectors, $\inner{\ub_w}{\vb_c}$;
the more relevant a context is, the larger we want their inner product
to be. We achieve this by learning a ranking model that is
parametrized by $\Ub$ and $\Vb$. If we sort the set of contexts
$\Ccal$ for a given word $w$ in terms of each context's inner product
score with the word, the rank of a specific context $c$ in this list
can be written as \cite{UsuBufGal09}:
\begin{align}
  \rank\rbr{w,c} &=\mkern-15mu\sum_{c' \in \Ccal \setminus \cbr{c}}
  I\rbr{\inner{\ub_{w}}{\vb_{c}} - \inner{\ub_{w}} {\vb_{c'}} \leq 0} \nonumber \\
  &=\mkern-15mu
  \sum_{c' \in \Ccal \setminus \cbr{c}} I\rbr{\inner{\ub_{w}}{\vb_{c} - \vb_{c'}} \leq 0},
  \label{eq:rankdef}
\end{align}
where $I(x \leq 0)$ is a 0-1 loss function which is 1 if $x \leq 0$
and $0$ otherwise. Since $I(x \leq 0)$ is a discontinuous function, we
follow the popular strategy in machine learning which replaces the 0-1
loss by its convex upper bound $\ell(\cdot)$, where $\ell(\cdot)$ can
be any popular loss function for binary classification such as the
hinge loss $\ell(x) = \max\rbr{0, 1 - x}$ or the logistic loss
$\ell(x) = \log_2\rbr{1 + 2^{-x}}$ \cite{JorBarMcA06}.  This enables
us to construct the following convex upper bound on the rank:
\begin{align}
  \rank\rbr{w,c} \mkern-3mu \leq \mkern-3mu \overline{\rank}\rbr{w, c}\mkern-5mu =\mkern-20mu 
  \sum_{c' \in \Ccal
  \setminus \cbr{c}}\mkern-18mu \ell\mkern-3mu\rbr{\mkern-2mu
  \inner{\ub_{w}}{\vb_{c}\mkern-5mu -\mkern-5mu \vb_{c'}}\mkern-2mu}
  \label{eq:rankub}
\end{align}

\begin{figure*}[htb]\vspace{-0.2cm}
  \begin{center}
    \subfigure{\includegraphics[width=2.0in]{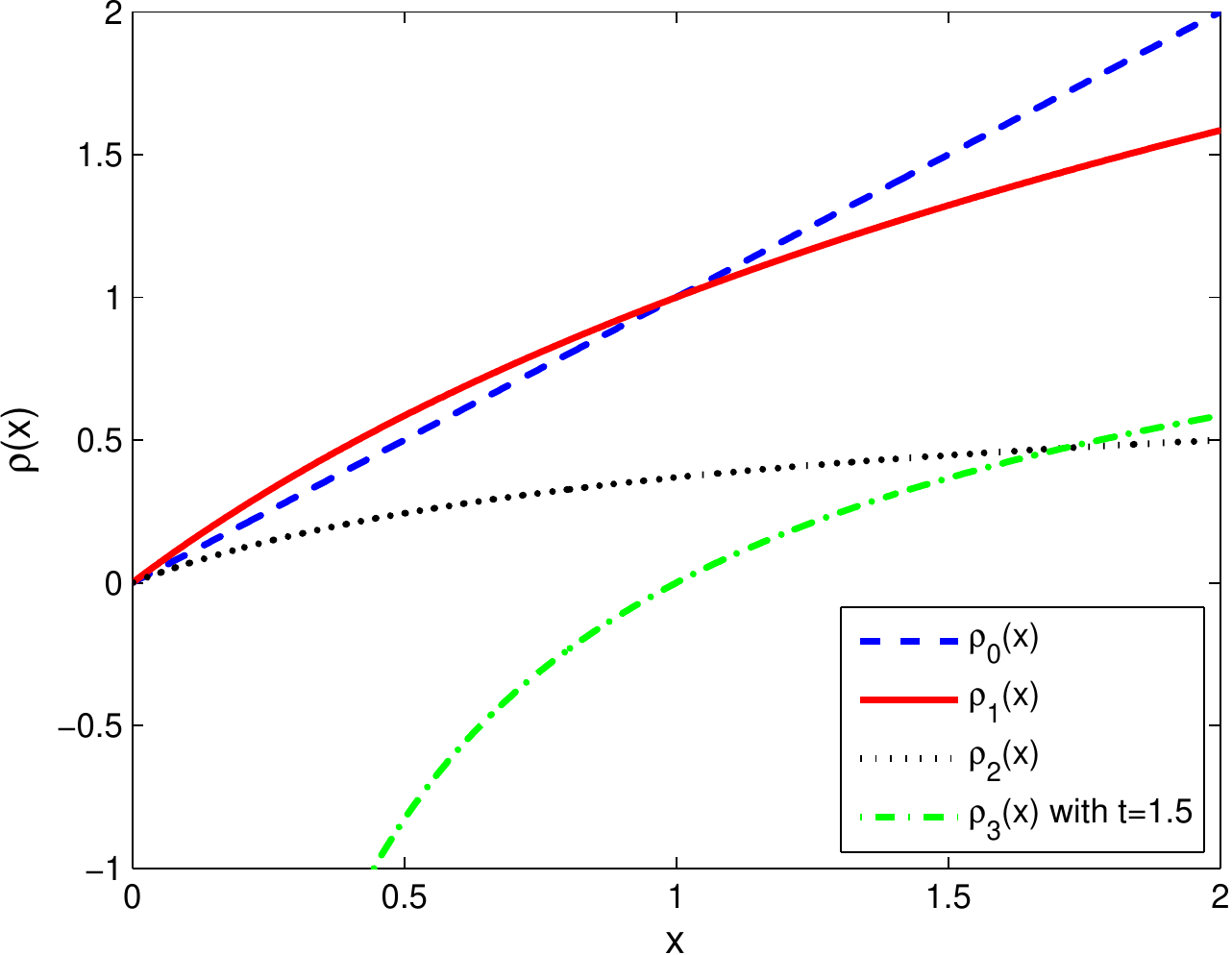}}\hspace{0.6in}
    \subfigure{\includegraphics[width=2.0in]{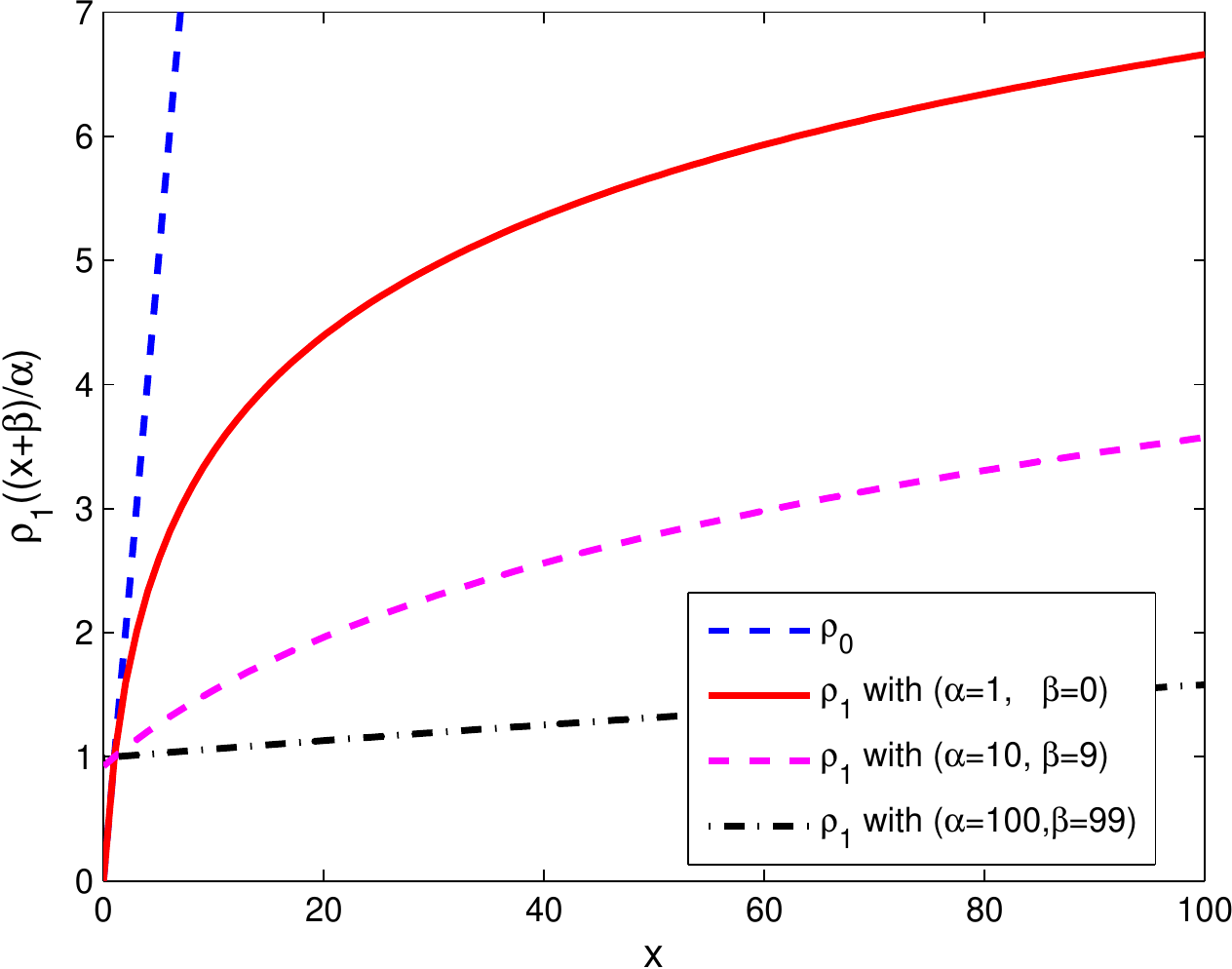}}
  \end{center}\vspace{-0.4cm}
  \caption{(a) Visualizing different ranking loss functions $\rho(x)$ as defined in Eqs.~(\ref{eq:id}--\ref{eq:logt}); the lower part of $\rho_3(x)$ is truncated in order to visualize the other functions better. (b) Visualizing $\rho_1((x+\beta)/\alpha)$ with different $\alpha$ and $\beta$; $\rho_0$ is included to illustrate the dramatic scale differences between $\rho_0$ and $\rho_1$.} 
  \label{fig:loss_func}
  \vspace{-0.4cm}
\end{figure*}

It is certainly desirable that the ranking model positions relevant
contexts at the top of the list; this motivates us to
write the objective function to minimize as:
\begin{align}
  J\rbr{\Ub, \Vb}\mkern-5mu :=\mkern-10mu \sum_{w \in \Wcal} \sum_{c \in \Omega_{w}} \mkern-8mu
  r_{w,c}\mkern-3mu \cdot\mkern-3mu \rho\rbr{\mkern-3mu\frac{\overline{\rank}\rbr{w,c}\mkern-3mu +\mkern-3mu \beta}{\alpha}
  \mkern-3mu}
  \label{eq:objective}
\end{align}
where $r_{w,c}$ is the weight between word $w$ and context $c$
quantifying the association between them, $\rho(\cdot)$ is a
monotonically increasing and concave ranking loss function that measures goodness of
a rank, and $\alpha>0$, $\beta>0$ are the hyperparameters of the model
whose role will be discussed later. Following \newcite{PenSocMan14}, we
use
\begin{align}
  \label{eq:wt-func}
  r_{w, c} =
  \begin{cases}
    \rbr{{X_{w,c}/x_{max}}}^\epsilon
    & \text{ if } X_{w,c} < x_{max} \\
    \hspace{0.23in}1 & \text{ otherwise},
  \end{cases}
\end{align}
where we set $x_{max}=100$ and $\epsilon=0.75$ in our experiments. That
is, we assign larger weights (with a saturation) to contexts that appear
more often with the word of interest, and vice-versa.  For the ranking
loss function $\rho(\cdot)$, on the other hand, we consider the class of
monotonically increasing and concave functions. While monotonicity is a
natural requirement, we argue that concavity is also important so that
the derivative of $\rho$ is always non-increasing; this implies that the
ranking loss to be the most sensitive at the top of the list (where the
rank is small) and becomes less sensitive at the lower end of the list
(where the rank is high). Intuitively this is desirable, because we are interested in a small
number of relevant contexts which frequently co-occur with a given word,
and thus are willing to tolerate errors on infrequent contexts\footnote{This is similar to 
the attention mechanism found in human visual system that is able to 
focus on a certain region of an image with ``high
resolution" while perceiving the surrounding image in ``low resolution" 
\cite{LarHin10,MniHeeGra14}.}. Meanwhile, this
insensitivity at the bottom of the list makes the model robust to noise
in the data either due to grammatical errors or unconventional use of
language. Therefore, a single ranking loss function $\rho(\cdot)$ serves two different purposes at two ends of the curve (see the example plots of $\rho$ in Figure~\ref{fig:loss_func}); while the left hand side of the curve encourages ``high resolution" on most relevant words, the right hand side becomes less sensitive (with ``low resolution") to infrequent and possibly noisy words\footnote{Due to the linearity of $\rho_0(x)
\mkern-3mu=\mkern-3mux$, this ranking loss doesn't have the benefit
of attention mechanism and robustness to noise since it treats all ranking errors uniformly.}. As we will demonstrate in our experiments, this is a fundamental attribute (in addition to the ranking nature) of our method that contributes its superior performance as compared to the state-of-the-arts when the training set is limited (\emph{i.e.}, sparse and noisy). 

What are interesting loss functions that can be used for
$\rho\rbr{\cdot}$? Here are four possible alternatives, all of which
have a natural interpretation (see the plots of all four $\rho$ functions in Figure~\ref{fig:loss_func}(a) and the related work in Sec.~\ref{sec:RelatedWork} for a discussion).
\begin{align}
  \label{eq:id}
  \rho_0\rbr{x} &:=x& \text{(identity)}\\
  \label{eq:log}
  \rho_1\rbr{x} &:=\log_{2}\rbr{1+x}& \text{(logarithm)}\\
  \label{eq:dcg}
  \rho_2\rbr{x} &:=1\mkern-3mu-\mkern-3mu\frac{1}{\log_2(2+x)}& \text{(negative DCG)}\\
  \label{eq:logt}
  \rho_3\rbr{x} &:=\frac{x^{1-t} - 1} {1-t}& (\log_{t} \text{ with } t\neq 1)
\end{align}
We will explore the performance of each of these variants in our
experiments.  For now, we turn our attention to efficient stochastic
optimization of the objective function \eqref{eq:objective}.

\subsection{Stochastic Optimization}
\label{sec:StochOptim}

Plugging \eqref{eq:rankub} into \eqref{eq:objective}, and replacing
$\sum_{w \in \Wcal} \sum_{c \in \Omega_{w}}$ by
$\sum_{\rbr{w,c} \in \Omega}$, the objective function becomes:
\begin{align}
  \label{eq:rewrite}
  J\rbr{\Ub, \Vb} &= \sum_{\rbr{w,c} \in \Omega} r_{w,c} \cdot \nonumber\\ 
  &\mkern-13mu\rho\rbr{\mkern-3mu
  \frac{\sum_{c' \in \Ccal
  \setminus \cbr{c}}\mkern-3mu \ell\rbr{
  \inner{\ub_{w}}{\vb_{c}\mkern-3mu -\mkern-3mu \vb_{c'}}}\mkern-3mu +\mkern-3mu \beta
  }{\alpha}\mkern-3mu}.
\end{align}
This function contains summations over $\Omega$ and $\Ccal$, both of
which are expensive to compute for a large corpus. Although stochastic
gradient descent (SGD) \cite{BotBou11} can be used to replace the
summation over $\Omega$ by random sampling, the summation over $\Ccal$
cannot be avoided unless $\rho(\cdot)$ is a linear function. To work
around this problem, we propose to optimize a linearized upper bound
of the objective function obtained through a first-order Taylor
approximation. Observe that due to the concavity of $\rho(\cdot)$, we have
\begin{align}
  \rho(x)
  \leq \rho\rbr{\xi^{-1}} + \rho'\rbr{\xi^{-1}} \cdot \rbr{x-\xi^{-1}}
  \label{eq:concavity}
\end{align}
for any $x$ and $\xi \neq 0$. Moreover, the bound is tight when
$\xi = x^{-1}$.  This motivates us to introduce a set of auxiliary
parameters $\Xi := \cbr{\xi_{w,c}}_{(w,c) \in \Omega}$ and define the
following upper bound of $J\rbr{\Ub, \Vb}$:
\begin{align}
  &\overline{J}\rbr{\Ub, \Vb, \Xi} := \sum_{\rbr{w,c} \in
  \Omega} 
  r_{w, c} \cdot 
  \Bigg\{
  \rho(\xi^{-1}_{wc}) + 
  \rho'(\xi^{-1}_{wc}) \mkern3mu\cdot \nonumber\\
  &\bigg(\mkern-3mu
  \alpha\mkern-3mu^{-1}\beta \mkern-3mu + \mkern-3mu \alpha\mkern-3mu^{-1}\mkern-15mu
  \sum_{c' \in
  \Ccal \setminus \cbr{c}} \mkern-15mu \ell\rbr{\inner{\ub_{w}}{\vb_{c}\mkern-3mu -\mkern-3mu \vb_{c'}}} \mkern-3mu
  - \mkern-3mu\xi^{-1}_{w,c}
  \bigg)
 \Bigg\}.
  \label{eq:obj_ubd}
\end{align}
Note that $J\rbr{\Ub, \Vb} \leq \overline{J}\rbr{\Ub, \Vb, \Xi}$ for any
$\Xi$, due to \eqref{eq:concavity}\footnote{When $\rho\mkern-3mu=\mkern-3mu \rho_{0}$, one can simply set the auxiliary variables $\xi_{w,c}\mkern-3mu=\mkern-3mu1$ because $\rho_0$ is
  already a linear function.}. Also, minimizing \eqref{eq:obj_ubd}
yields the same $\Ub$ and $\Vb$ as minimizing \eqref{eq:rewrite}. To see
this, suppose $\hat{\Ub} := \cbr{\hat{\ub}_w}_{w \in \Wcal}$ and
$\hat{\Vb} := \cbr{\hat{\vb}_c}_{c \in \Ccal}$ minimizes
\eqref{eq:rewrite}. Then, by letting
$\hat{\Xi} := \cbr{\hat{\xi}_{w,c}}_{(w,c) \in \Omega}$ where
\begin{align}
  \hat{\xi}_{w,c} & = \frac{\alpha}{ 
                    \sum_{c' \in \Ccal \setminus \cbr{c}}
                    \ell\rbr{ \inner{\hat{\ub}_{w}}{\hat{\vb}_{c} - \hat{\vb}_{c'}}} + \beta}, 
  \label{eq:xi_update}
\end{align}
we have
$\overline{J}\rbr{\hat{\Ub}, \hat{\Vb}, \hat{\Xi}} = J\rbr{\hat{\Ub},
  \hat{\Vb}}$.
Therefore, it suffices to optimize \eqref{eq:obj_ubd}. However, unlike
\eqref{eq:rewrite}, \eqref{eq:obj_ubd} admits an efficient SGD
algorithm. To see this, rewrite \eqref{eq:obj_ubd} as
\begin{align}
  &\overline{J}\rbr{\Ub, \Vb, \Xi} \mkern-3mu:= 
  \mkern-20mu
  \sum_{\rbr{w,c,c'} } 
  \mkern-12mu
r_{w,c} 
  \mkern-5mu
\cdot 
    \mkern-5mu
  \Bigg(\mkern-3mu
  \frac{\rho(\xi^{-1}_{w,c})\mkern-3mu+\mkern-3mu\rho'(\xi^{-1}_{w,c})\mkern-3mu\cdot\mkern-3mu(\alpha\mkern-3mu^{-1}\beta\mkern-3mu-\mkern-3mu\xi^{-1}_{w,c})}{\abr{\Ccal} - 1} 
  \nonumber \\ &\mkern50mu+ 
  \frac{1}{\alpha}\rho'(\xi^{-1}_{w,c}) \cdot 
  \ell\rbr{\inner{\ub_{w}}{\vb_{c} - \vb_{c'}}}
  \Bigg)
  ,
  \label{eq:unrolled}
\end{align}
where $\rbr{w, c, c'} \in \Omega \times \rbr{\Ccal \setminus \cbr{c}}$.  Then,
it can be seen that if we sample uniformly from
$(w,c) \in \Omega$ and $c' \in \Ccal \setminus \cbr{c}$, then
$ j(w,c,c') :=$
\begin{align}
\mkern-8mu \abr{\Omega} \mkern-3mu \cdot \mkern-3mu \rbr{\abr{\Ccal}\mkern-3mu-\mkern-3mu 1}
  \cdot \mkern3mu &r_{w, c} \mkern-3mu \cdot \mkern-3mu
  \Bigg(\mkern-3mu
  \frac{\rho(\xi^{-1}_{w,c})\mkern-3mu +\mkern-3mu
  \rho'(\xi^{-1}_{w,c})\mkern-3mu\cdot\mkern-3mu(\alpha\mkern-3mu^{-1}\beta\mkern-3mu -\mkern-3mu
  \xi^{-1}_{w,c})}{\abr{\Ccal} - 1}  
  \nonumber\\ &\mkern-40mu+ 
  \frac{1}{\alpha}\rho'(\xi^{-1}_{w,c}) \mkern-3mu\cdot \mkern-3mu
  \ell\rbr{\inner{\ub_{w}}{\vb_{c} - \vb_{c'}}}
  \Bigg)
  ,
  \label{eq:def_jwc}
\end{align}
which does not contain any expensive summations and is an unbiased
estimator of \eqref{eq:unrolled}, i.e.,
$\EE\sbr{j(w,c,c')} = \overline{J}\rbr{\Ub, \Vb, \Xi}$. On the
other hand, one can optimize $\xi_{w,c}$ exactly by using
\eqref{eq:xi_update}. Putting everything together yields a stochastic optimization algorithm \wrr, which can be
specialized to a variety of ranking loss functions $\rho(\cdot)$ with
weights $r_{w,c}$ (e.g., DCG (Discounted Cumulative Gain)
\cite{ManRagSch08} is one of many possible
instantiations). Algorithm~\ref{alg:twostage} contains detailed
pseudo-code. It can be seen that the algorithm is divided into two
stages: a stage that updates $(\Ub, \Vb)$ and another that updates
$\Xi$. Note that the time complexity of the first stage is
$\Ocal(\abr{\Omega})$ since the cost of each update in Lines~8--10 %\ref{alg:up1}--\ref{alg:up3} 
is independent of the size of the
corpus. On the other hand, the time complexity of updating $\Xi$ in
Line~15 %\ref{alg:xi_update} 
is $\Ocal(\abr{\Omega} \abr{\Ccal})$, which can be
expensive. To amortize this cost, we employ two tricks: we only update
$\Xi$ after a few iterations of $\Ub$ and $\Vb$ update, and we
exploit the fact that the most computationally expensive operation in  
\eqref{eq:xi_update} 
involves a matrix and matrix multiplication which can be calculated
efficiently via the SGEMM routine in BLAS~\cite{DonCroDufHam90}. 

\begin{algorithm}
  \begin{algorithmic}[1]
    \STATE {$\eta$: step size}
    \REPEAT
    \STATE {\texttt{// Stage 1: Update $\Ub$ and $\Vb$}}
    \REPEAT
    \STATE {Sample $(w,c)$ uniformly from $\Omega$}
    \STATE {Sample $c'$ uniformly from $\Ccal \setminus \cbr{c}$}
    \STATE {\texttt{// following three updates are executed simultaneously}}
    \STATE {$\ub_w\mkern-3mu \leftarrow\mkern-3mu \ub_w - \eta \cdot r_{w,c} \cdot \rho'(\xi^{-1}_{w,c})
      \cdot \ell'\rbr{\inner{\ub_{w}}{\vb_{c}\mkern-3mu -\mkern-3mu \vb_{c'}}} \cdot
      (\vb_c\mkern-3mu -\mkern-3mu \vb_{c'})$} \label{alg:up1}
    \STATE {$\vb_c\mkern-3mu \leftarrow\mkern-3mu \vb_c - \eta \cdot r_{w,c} \cdot \rho'(\xi^{-1}_{w,c})
      \cdot \ell'\rbr{\inner{\ub_{w}}{\vb_{c}\mkern-3mu -\mkern-3mu \vb_{c'}}} \cdot
      \ub_w$} \label{alg:up2}
    %\vspace{-0.2cm}
    \STATE {$\vb_{c'}\mkern-3mu \leftarrow\mkern-3mu \vb_{c'} + \eta \cdot r_{w,c} \cdot \rho'(\xi^{-1}_{w,c})
      \cdot \ell'\rbr{\inner{\ub_{w}}{\vb_{c}\mkern-3mu -\mkern-3mu \vb_{c'}}} \cdot
      \ub_w$}     \label{alg:up3}
    %\vspace{-0.2cm}
    \UNTIL {$\Ub$ and $\Vb$ are converged}
        \STATE {\texttt{// Stage 2: Update $\Xi$}}
    \FOR {$w \in \Wcal$}
    \FOR {$c \in \Ccal$}
    \STATE $\mkern-13mu\xi_{w,c}\mkern-3mu =\mkern-3mu \alpha /\mkern-3mu \rbr{\mkern-3mu\sum_{c'\in \Ccal \setminus \cbr{c}}
    \ell\rbr{\inner{\ub_{w}}{\vb_{c}\mkern-3mu -\mkern-3mu \vb_{c'}}}\mkern-3mu +\mkern-3mu \beta}$ \label{alg:xi_update}
    \vspace{-0.3cm}
    \ENDFOR
    \ENDFOR
    \UNTIL {$\Ub$, $\Vb$ and $\Xi$ are converged}
  \end{algorithmic}
  \caption{\wrr algorithm.}
  \label{alg:twostage}
\end{algorithm}

\subsection{Parallelization}
\label{sec:Parallelization}
The updates in Lines~8--10 %\ref{alg:up1}--\ref{alg:up3} 
have one remarkable
property: To update $\ub_{w}$, $\vb_{c}$ and $\vb_{c'}$, we only need to
\emph{read} the variables $\ub_{w}$, $\vb_{c}$, $\vb_{c'}$ and
$\xi_{w, c}$. What this means is that updates to another triplet of
variables $\ub_{\hat{w}}$, $\vb_{\hat{c}}$ and $\vb_{\hat{c'}}$ can be
performed independently. This observation is the key to developing a
parallel optimization strategy, by distributing the computation of the
updates among multiple processors. Due to lack of space, details
including pseudo-code are relegated to
the supplementary material. %~\ref{sec:ParallelAlgorithm}. 

\subsection{Interpreting of $\alpha$ and $\beta$}
\label{sec:Interpralphabeta}

The update \eqref{eq:xi_update} indicates that $\xi^{-1}_{w,c}$ is
proportional to $\overline{\rank}\rbr{w,c}$. On the other hand, one can
observe that the loss function $\ell\rbr{\cdot}$ in \eqref{eq:def_jwc}
is weighted by a $\rho'\rbr{\xi^{-1}_{w,c}}$ term. Since
$\rho\rbr{\cdot}$ is concave, its gradient $\rho'\rbr{\cdot}$
is monotonically non-increasing ~\cite{Rockafellar70}. Consequently, when
$\overline{\rank}\rbr{w,c}$ and hence $\xi^{-1}_{w,c}$ is large,
$\rho'\rbr{\xi^{-1}_{w,c}}$ is small. In other words, the loss function
``gives up'' on contexts with high ranks in order to focus its
attention on top of the list. The rate at which the algorithm gives up
is determined by the hyperparameters $\alpha$ and $\beta$. For the illustration of this effect, see the example plots of $\rho_1$ with different $\alpha$ and $\beta$ in Figure~\ref{fig:loss_func}(b). Intuitively,
$\alpha$ can be viewed as a \emph{scale} parameter while $\beta$ can be
viewed as an \emph{offset} parameter. An equivalent interpretation is
that by choosing different values of $\alpha$ and $\beta$ one can modify
the behavior of the ranking loss $\rho\rbr{\cdot}$ in a problem
dependent fashion. In our experiments, we found that a common setting of
$\alpha\mkern-3mu=\mkern-3mu 1$ and $\beta\mkern-3mu=\mkern-3mu 0$ often yields uncompetitive performance, while setting $\alpha\mkern-3mu =\mkern-3mu 100$ and $\beta\mkern-3mu =\mkern-3mu 99$ generally gives good results.

%\section{Related Work}
%%\label{sec:RelatedWork}
%Our work sits at the intersection of word embedding and ranking optimization. As we discussed
%above, it's also related to the attention mechanism widely used in deep learning. We
%therefore review related work along these three axes. Due to the space constraint, detailed discussions are relegated to Appendix~\ref{sec:RelatedWork}.

\section{Related Work}
\label{sec:RelatedWork}
Our work sits at the intersection of word embedding and ranking optimization. As we discussed in Sec.~\ref{sec:RankingModel} and Sec.~\ref{sec:Interpralphabeta}, it's also related to the attention mechanism widely used in deep learning. We therefore review the related work along these three axes.

\paragraph{Word Embedding.}
We already discussed some related work (\wv and \gl) on word embedding in the introduction. Essentially, \wv and \gl derive word representations by modeling a transformation (PMI or log) of $X_{w,c}$ directly, while \wrr learns word representations via robust ranking. Besides these state-of-the-art techniques, a few ranking-based approaches have been proposed for word embedding recently, e.g., \cite{ColWes08,VilMcc15,LiuJiaWei15}. However, all of them adopt a pair-wise binary classification approach with a linear ranking loss $\rho_0$. For example, \cite{ColWes08,VilMcc15} employ a hinge loss on positive/negative word pairs to learn word representations and $\rho_0$ is used \emph{implicitly} to evaluate ranking losses. As we discussed in Sec.~\ref{sec:RankingModel}, $\rho_0$ has no benefit of the attention mechanism and robustness to noise since its linearity treats all the ranking errors uniformly; empirically, sub-optimal performances are often observed with $\rho_0$ in our experiments. More recently, by extending the Skip-Gram model of \wv, \newcite{LiuJiaWei15} incorporates additional pair-wise constraints induced from 3rd-party knowledge bases, such as WordNet, and learns word representations jointly. In contrast, \wrr is a fully ranking-based approach without using any additional data source for training. %Furthermore, all the aforementioned word embedding methods are online learning algorithms, while \wrr and \gl are matrix factorization based, in which a co-occurrence matrix is built explicitly to start from.

\paragraph{Robust Ranking.} The second line of work that is very relevant 
to \wrr is that of ranking objective (\ref{eq:objective}). The use of score functions
$\inner{\ub_{w}}{\vb_{c}}$ for ranking is inspired by the latent
collaborative retrieval framework of \newcite{WesWanWeiBer12}. Writing
the rank as a sum of indicator functions \eqref{eq:rankdef}, and upper
bounding it via a convex loss \eqref{eq:rankub} is due to
\newcite{UsuBufGal09}. Using $\rho_{0}\rbr{\cdot}$ \eqref{eq:id}
corresponds to the well-known pairwise ranking loss (see \eg,
\cite{LeeLin13}). On the other hand, \newcite{YunRamVis14} observed that
if they set $\rho\mkern-3mu =\mkern-3mu \rho_{2}$ as in \eqref{eq:dcg}, then
$-J\rbr{\Ub, \Vb}$ corresponds to the DCG (Discounted Cumulative Gain), 
one of the most popular ranking metrics used in web search ranking \cite{ManRagSch08}. 
In their RobiRank algorithm they proposed the use of $\rho = \rho_{1}$
\eqref{eq:log}, which they considered to be a special function for which
one can derive an efficient stochastic optimization procedure. However,
as we showed in this paper, the general class of monotonically
increasing concave functions can be handled efficiently. Another
important difference of our approach is the hyperparameters $\alpha$
and $\beta$, which we use to modify the behavior of $\rho$, and which we
find are critical to achieve good empirical results. \newcite{DinVis10}
proposed the use of $\rho\mkern-3mu =\mkern-3mu \log_{t}$ in the context of robust
\emph{binary} classification, while here we are concerned with
ranking, and our formulation is very general and applies to a variety of ranking
losses $\rho\rbr{\cdot}$ with weights $r_{w,c}$. Optimizing over $\Ub$
and $\Vb$ by distributing the computation across processors is inspired
by work on distributed stochastic gradient for matrix factorization
\cite{GemNijHaaSis11}.

\paragraph{Attention.}
Attention is one of the most important advancements in deep learning in recent years \cite{LarHin10}, and is now widely used in state-of-the-art image recognition and machine translation systems \cite{MniHeeGra14,BahChoBen15}. Recently, attention has also been applied to the domain of word embedding. For example, under the intuition that not all contexts are created equal, \newcite{WanLinTsv15} assign an importance weight to each word type at each context position and learn an attention-based Continuous Bag-Of-Words (CBOW) model. Similarly, within a ranking framework, \wrr expresses the context importance by introducing the auxiliary variable $\xi_{w,c}$, which ``gives up" on contexts with high ranks in order to focus its attention on top of the list.

\section{Experiments} 
\label{sec:Experiments}
In our experiments, we first evaluate the impact of the weight
$r_{w,c}$ and the ranking loss function $\rho(\cdot)$ on the test performance
using a small dataset. We then pick the best performing model and
compare it against \wv \cite{MikSutCheCoretal13} and \gl
\cite{PenSocMan14}. We closely follow the framework of
\newcite{LevGolDag15} to set up a careful and fair comparison of the three
methods. Our code is publicly available at \url{https://bitbucket.org/shihaoji/wordrank}.

\begin{table*}[htb]   \vspace{-0.3cm}
  \caption {Parameter settings used in the experiments.} \vspace{-0.1cm}
  \label{tab:paramters}
  \begin{center}
    \begin{threeparttable}
      \begin{tabular}{|l|c|c|c|c|c|c|c|c|c|}
        \hline Corpus Size      & 17M$^*$  & 32M  & 64M  & 128M & 256M & 512M & 1.0B & 1.6B & 7.2B\\
        \hline Vocabulary Size $\abr{\Wcal}$ & 71K  & 100K & 100K & 200K & 200K & 300K & 300K & 400K & 620K\\
        \hline Window Size $win$    & 15   & 15   & 15   & 10   & 10   & 10   & 10   & 10   & 10\\
        \hline Dimension $k$      & 100  & 100  & 100  & 200  & 200  & 300  & 300  & 300  & 300\\
        \hline
      \end{tabular}%\vspace{-0.1cm}
      \begin{tablenotes}
        \scriptsize\item * This is the Text8 dataset from
        \url{http://mattmahoney.net/dc/text8.zip}, which is widely used
        for word embedding demo. 
      \end{tablenotes}
    \end{threeparttable}
  \end{center}\vspace{-0.4cm}
\end{table*}

\begin{table*}[htb]\vspace{-0.1cm}
  \caption {Performance of different $\rho$ functions on Text8 dataset
    with 17M tokens.} \vspace{-0.1cm}
  \label{tab:small_exp}
  \begin{center}
    \begin{tabular}{|l|c|c|c|c|c|c|c|c|c|}
      \hline Task                & Robi             & \multicolumn{2}{c|}{$\rho_{0}$} & \multicolumn{2}{c|}{$\rho_{1}$} & \multicolumn{2}{c|}{$\rho_2$} & \multicolumn{2}{c|}{$\rho_{3}$}        \\
                  \cline{3 - 10} &                  & off                             & on                              & off                           & on               & off  &  on   & off & on \\
      \hline Similarity          & 41.2             & 69.0                            & \underline{71.0}                            & 66.7                          & 70.4             & 66.8 & 70.8  & 68.1 & 68.0 \\
      \hline Analogy             & 22.7             & 24.9                            & 31.9                            & 34.3                          & \underline{44.5}
                                 & 32.3              & 40.4                            & 33.6                           & 42.9                                                                    \\
      \hline
    \end{tabular}%\vspace{-0.3cm}
  \end{center}\vspace{-0.3cm}
\end{table*}

\paragraph{Training Corpus}
Models are trained on a combined corpus of 7.2 billion tokens, which
consists of the 2015 Wikipedia dump with 1.6 billion tokens, the WMT14
News
Crawl\footnote{\url{http://www.statmt.org/wmt14/translation-task.html}}
with 1.7 billion tokens, the ``One Billion Word Language Modeling
Benchmark''\footnote{\url{http://www.statmt.org/lm-benchmark}} with
almost 1 billion tokens, and UMBC webbase
corpus\footnote{\url{http://ebiquity.umbc.edu/resource/html/id/351}}
with around 3 billion tokens. 
%The Wikipedia dump is in XML format and
%the article content needs to be parsed from wiki markups, while the
%other corpora are already in plain text format.
The pre-processing pipeline breaks the paragraphs into sentences, tokenizes and lowercases 
each corpus with the Stanford tokenizer. We further clean up the dataset 
by removing non-ASCII characters and punctuation, and discard sentences
that are shorter than 3 tokens or longer than 500 tokens. In the end, we 
obtain a dataset of 7.2 billion tokens, with the first 1.6 billion tokens 
from Wikipedia. When we want to experiment with a smaller corpus, we extract
a subset which contains the specified number of tokens.

\paragraph{Co-occurrence matrix construction}

We use the \gl code to construct the co-occurrence matrix $X$, and the
same matrix is used to train \gl and \wrr models.  When constructing
$X$, we must choose the size of the vocabulary, the context window and
whether to distinguish left context from right context. We follow the
findings and design choices of \gl and use a symmetric window of size
$win$ with a decreasing weighting function, so that word pairs that are
$d$ words apart contribute $1/d$ to the total count. Specifically, when
the corpus is small (\eg, 17M, 32M, 64M) we let $win\mkern-3mu =\mkern-3mu 15$ and for larger
corpora we let $win\mkern-3mu =\mkern-3mu 10$. The larger window size alleviates the data
sparsity issue for small corpus at the expense of adding more noise to
$X$. The parameter settings used in our experiments are summarized in
Table~\ref{tab:paramters}. 

\paragraph{Using the trained model}

It has been shown by \newcite{PenSocMan14} that combining the $\ub_w$ and
$\vb_c$ vectors with equal weights gives a small boost in
performance. This vector combination was originally motivated as an
ensemble method \cite{PenSocMan14}, and later \newcite{LevGolDag15}
provided a different interpretation of its effect on the cosine
similarity function, and show that adding context vectors effectively
adds first-order similarity terms to the second-order similarity
function. In our experiments, we find that vector combination boosts the
performance in word analogy task when training set is small, but when
dataset is large enough (\eg, 7.2 billion tokens), vector combination
doesn't help anymore. More interestingly, for the word similarity
task, we find that vector combination is detrimental in all the cases,
sometimes even substantially\footnote{This is possible since we optimize
  a ranking loss: the absolute scores don't matter as long as they yield
  an ordered list correctly. Thus, \wrr's $\ub_w$ and $\vb_c$ are
  less comparable to each other than those generated by \gl, which
  employs a point-wise $L_2$ loss.}.
Therefore, we will always use $\ub_w$ on word similarity task, and use
$\ub_w+\vb_c$ on word analogy task unless otherwise noted.

\subsection{Evaluation}
\label{sec:Evaluation}

\paragraph{Word Similarity} 

We use six datasets to evaluate word similarity: WS-353
\cite{FinGabMat02} partitioned into two subsets: WordSim Similarity and
WordSim Relatedness \cite{AgiAlfHal09}; MEN \cite{BruBolBar12};
Mechanical Turk \cite{RadAgiGab11}; Rare words \cite{LuoSocMan13}; and
SimLex-999 \cite{HilReiKor14}. They contain word pairs
together with human-assigned similarity judgments. The word
representations are evaluated by ranking the pairs according to their
cosine similarities, and measuring the Spearman's rank correlation
coefficient with the human judgments.

\paragraph{Word Analogies} 
For this task, we use the Google analogy dataset
\cite{MikCheCorDea13}. It contains 19544 word analogy questions,
partitioned into 8869 semantic and 10675 syntactic questions. 
%The semantic questions contain five types of semantic analogies, such as
%capital cities (\emph{Paris:France;Tokyo:?}), currency
%(\emph{USA:dollar;India:?}) or people (\emph{king:queen;man:?}). The
%syntactic questions contain nine types of analogies, such as plural
%nouns, opposite, or comparative, for example
%\emph{good:better;smart:?}. 
A question is correctly answered only if the
algorithm selects the word that is exactly the same as the correct word
in the question: synonyms are thus counted as mistakes. There are two
ways to answer these questions, namely, by using 3CosAdd or 3CosMul (see
\cite{LevGol14} for details). We will report scores by using 3CosAdd by
default, and indicate when 3CosMul gives better performance.

\paragraph{Handling questions with out-of-vocabulary words}

Some papers (\eg, \cite{LevGolDag15}) filter out questions with out-of-vocabulary words when reporting performance. By contrast, in our experiments if any word of a question is out of vocabulary, the corresponding question will be marked as unanswerable and will get a score of zero. This decision is
made so that when the size of vocabulary increases, the model
performance is still comparable across different experiments.

\subsection{The impact of $r_{w, c}$ and $\rho(\cdot)$}
\label{sec:impactwtrho}

In Sec. \ref{sec:RankingModel} we argued the need for adding weight
$r_{w,c}$ to ranking objective (\ref{eq:objective}), and we also presented 
our framework which can deal with a variety of ranking loss functions $\rho$. 
We now study the utility of these two ideas. We report
results on the 17 million token dataset in Table~\ref{tab:small_exp}. For the
similarity task, we use the WS-353 test set and for the analogy task we
use the Google analogy test set. The best scores for each task are
underlined. We set $t\mkern-3mu =\mkern-3mu 1.5$ for $\rho_{3}$.  ``Off'' means that we used
uniform weight $r_{w,c}\mkern-3mu =\mkern-3mu 1$, and ``on'' means that $r_{w,c}$ was set
as in \eqref{eq:wt-func}. For comparison, we also include the results
using RobiRank \cite{YunRamVis14}\footnote{We used the code provided by
  the authors at
  \url{https://bitbucket.org/d_ijk_stra/robirank}. Although related to
  RobiRank, we attribute the superior performance of \wrr to the use of weight $r_{w,c}$
  \eqref{eq:wt-func}, introduction of hyperparameters $\alpha$ and $\beta$,
  and many implementation details.}.

It can be seen from Table~\ref{tab:small_exp} that adding the weight $r_{w,c}$ improves performance in all the cases, especially on the word analogy task. Among the four $\rho$ functions, $\rho_{0}$ performs the best on the word similarity task but suffers notably on the analogy task, while $\rho_{1}\mkern-3mu =\mkern-3mu \log$ performs the best overall. Given these observations, which are consistent with the results on large scale datasets, in the experiments that follow we only report \wrr
with the best configuration, \ie, using $\rho_1$ with the weight $r_{w,c}$ as defined in \eqref{eq:wt-func}. 

\begin{figure*}[htb]\vspace{-0.3cm}
  \begin{center}
    \subfigure{\label{fig:scale_exp_a}\includegraphics[width=2.3in]{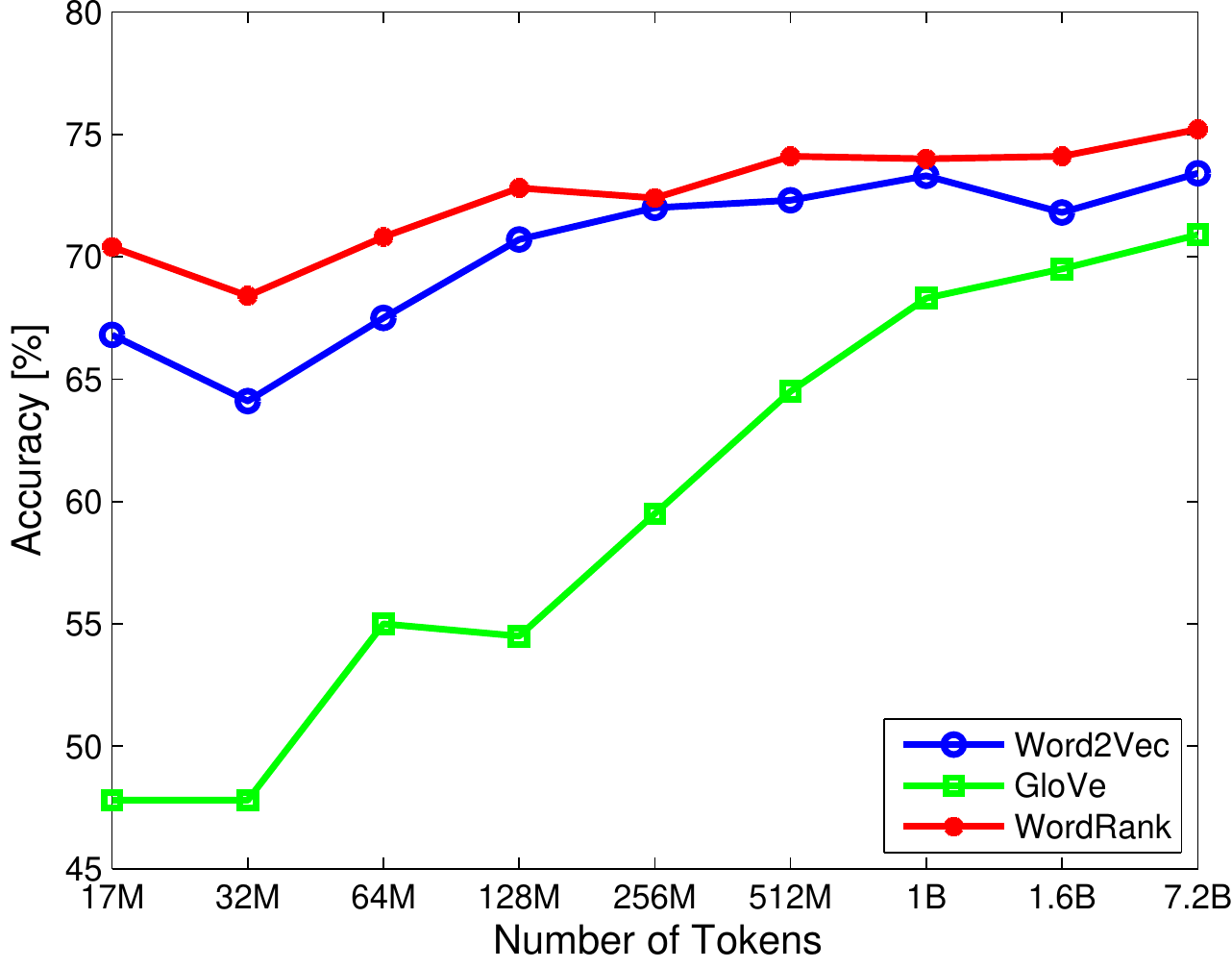}}\hspace{0.3in}
    \subfigure{\label{fig:scale_exp_b}\includegraphics[width=2.3in]{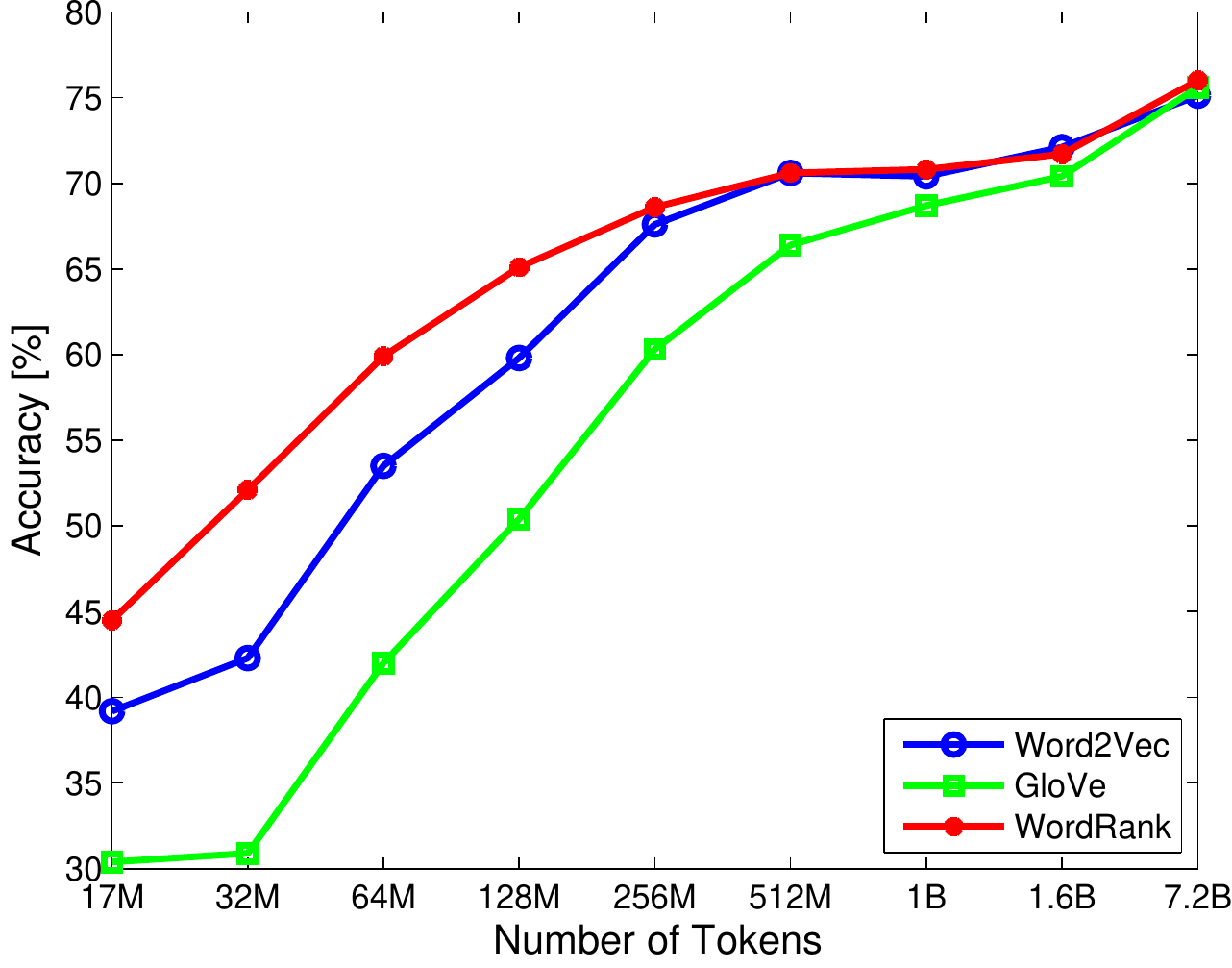}}
  \end{center}\vspace{-0.4cm}
  \caption{Performance evolution as a function of corpus size (a) on
    WS-353 word similarity benchmark; (b) on Google word analogy
    benchmark.}\label{fig:scale_exp}%\vspace{-0.3cm}
\end{figure*}

\begin{table*}[htb]\vspace{-0.2cm}
\caption {Performance of the best \wv, \gl and \wrr models, learned from 7.2 billion tokens, on six similarity tasks and Google semantic and syntactic subtasks. }\vspace{-0.1cm}
\label{tab:breakdown}
\begin{center}
\begin{tabular}{|l||c|c|c|c|c|c||c|c|}
\hline           & \multicolumn{6}{c||}{Word Similarity} & \multicolumn{2}{c|}{Word Analogy} \\
\cline{2-9}   Model & \small{WordSim} & \small{WordSim} & \small{Bruni et} & \small{Radinsky} & \small{Luong et} & \small{Hill et al.} & \small{Goog} & \small{Goog} \\
             & \small{Similarity} & \small{Relatedness} & \small{al. MEN} & \small{et al. MT} & \small{al. RW} & \small{SimLex} & \small{Sem.} & \small{Syn.} \\
\hline          \wv      & 73.9 & 60.9 & 75.4 & 66.4 & 45.5 & 36.6 & 78.8 & 72.0 \\
\hline          \gl     & 75.7 & 67.5 & \underline{78.8} & 69.7 & 43.6 & 41.6 & \underline{80.9} & 71.1 \\
\hline          \wrr  & \underline{79.4} & \underline{70.5} & 78.1 & \underline{73.5} & \underline{47.4} & \underline{43.5} & 78.4 & \underline{74.7} \\
\hline
\end{tabular}%\vspace{-0.3cm}
\end{center}\vspace{-0.3cm}
\end{table*}

\subsection{Comparison to state-of-the-arts}
\label{sec:Comparisonstateart}

In this section we compare the performance of \wrr with
\wv\footnote{\url{https://code.google.com/p/word2vec/}} and
\gl\footnote{\url{http://nlp.stanford.edu/projects/glove}}, by using the
code provided by the respective authors. For a fair comparison, \gl and
\wrr are given as input the same co-occurrence matrix $X$; this
eliminates differences in performance due to window size and other such
artifacts, and the same parameters are used to \wv. Moreover, the
embedding dimensions used for each of the three methods is the same (see
Table~\ref{tab:paramters}). With \wv, we train the Skip-Gram with
Negative Sampling (SGNS) model since it produces state-of-the-art
performance, and is widely used in the NLP community
\cite{MikSutCheCoretal13}. For \gl, we use the default parameters as
suggested by \cite{PenSocMan14}. The results are provided in
Figure~\ref{fig:scale_exp} (also see Table \ref{tab:scale_exp} in
the supplementary material %\ref{sec:AdditExperDeta} 
for additional details).

As can be seen, when the size of corpus increases, in general all
three algorithms improve their prediction accuracy on both tasks. This
is to be expected since a larger corpus typically produces better
statistics and less noise in the co-occurrence matrix $X$. When the
corpus size is small (\eg, 17M, 32M, 64M, 128M), \wrr yields the best
performance with significant margins among three, followed by \wv and \gl; 
when the size of corpus increases further, on the word analogy task \wv and \gl become very competitive to \wrr, and eventually perform neck-to-neck to each other (Figure~\ref{fig:scale_exp}(b)). This is consistent with
the findings of \cite{LevGolDag15} indicating that when the number of
tokens is large even simple algorithms can perform well. On the other hand, \wrr is dominant on the word similarity task for all the cases (Figure~\ref{fig:scale_exp}(a)) since it optimizes a ranking loss \emph{explicitly}, which aligns more naturally with the objective of word similarity than the other methods; with 17 million tokens our method performs almost 
as well as existing methods using 7.2 billion tokens on the word similarity benchmark.

As a side note, on a similar 1.6-billion-token Wikipedia corpus, our \wv
and \gl performance scores are somewhat better than or close to the
results reported by \newcite{PenSocMan14}; and our \wv and \gl scores on
the 7.2-billion-token dataset are close to what they reported on a
42-billion-token dataset. We believe this discrepancy is primary
due to the extra attention we paid to pre-process the Wikipedia
and other corpora.

To further evaluate the model performance on the word similarity/analogy tasks, we use the best performing models trained on the 7.2-billion-token corpus to predict on the six
word similarity datasets described in Sec.~\ref{sec:Evaluation}. Moreover, we breakdown the performance of the
models on the Google word analogy dataset into the semantic and
syntactic subtasks. Results are listed in Table~\ref{tab:breakdown}. As
can be seen, \wrr outperforms \wv and \gl on 5 of 6 similarity tasks,
and 1 of 2 Google analogy subtasks.

\section{Visualizing the results}
\label{sec:Visualizingresults}

\begin{figure}[htb]
\centering
\includegraphics[width=3.0in]{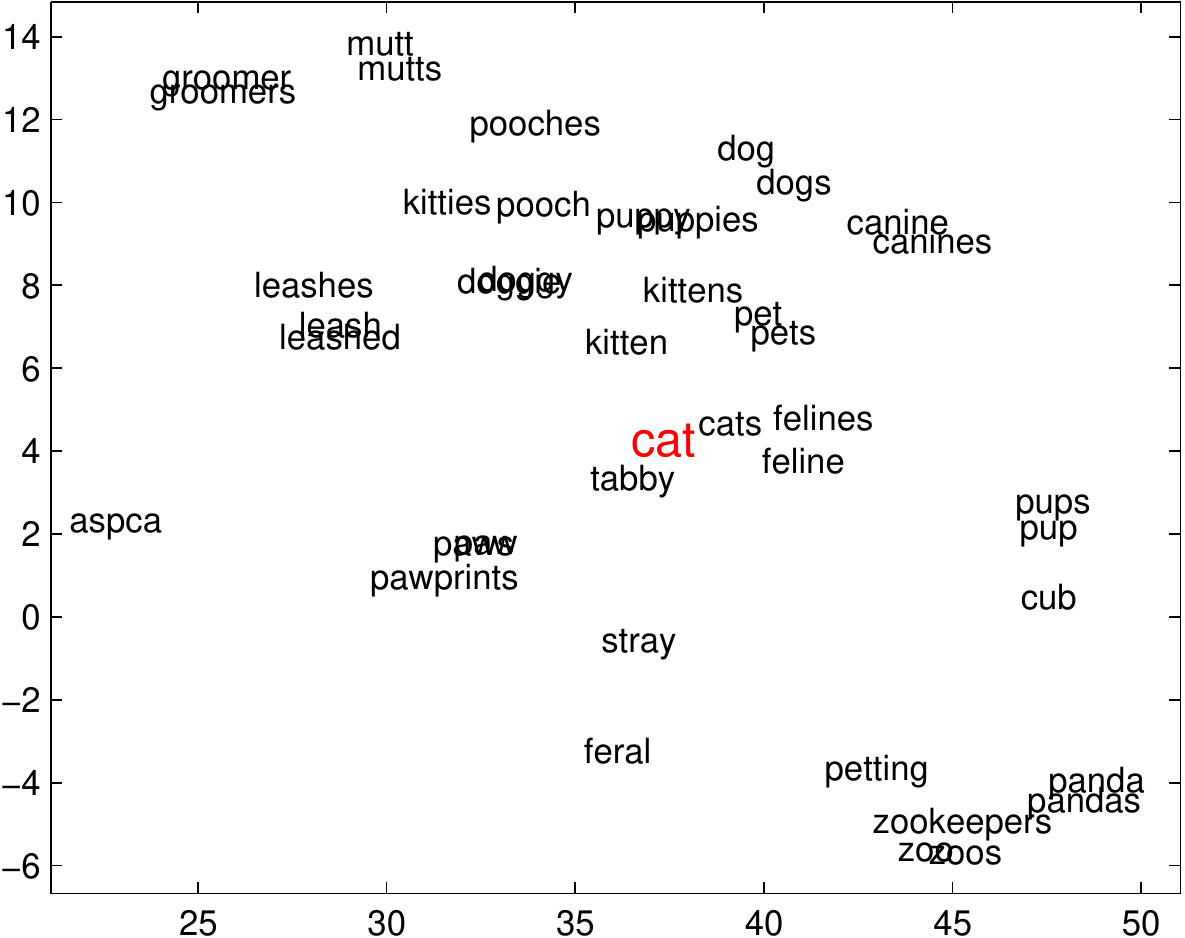} %\vspace{-0.2cm}
\caption{Nearest neighbors of ``cat'' found by projecting a 300d word
  embedding learned from \wrr onto a 2d space.}
\label{fig:cat}
%\vspace{-0.3cm}
\end{figure}

To understand whether \wrr produces \emph{syntatically} and
\emph{semantically} meaningful vector space, we did the following experiment:
we use the best performing model produced using 7.2 billion tokens, and compute the
nearest neighbors of the word ``cat''. We then visualize the words in
two dimensions by using t-SNE %, a well-known technique for dimensionality reduction
~\cite{MaaHin08}. As can be seen in
Figure~\ref{fig:cat}, our ranking-based model is indeed capable of capturing both
semantic (\eg, cat, feline, kitten, tabby) and syntactic (\eg, leash,
leashes, leashed) regularities of the English language.

\section{Conclusion}
\label{sec:Conclusion}
We proposed \wrr, a ranking-based approach, to learn word representations from large scale textual corpora. The most prominent difference between our method and the state-of-the-art techniques, such as \wv and \gl, is that \wrr learns word representations via a robust ranking model, while \wv and \gl typically model a transformation of co-occurrence count $X_{w,c}$ directly. Moreover, by a ranking loss function $\rho(\cdot)$, \wrr achieves its attention mechanism and robustness to noise naturally, which are usually lacking in other ranking-based approaches. These attributes significantly boost the performance of \wrr in the cases where training data are sparse and noisy. Our multi-node distributed implementation of \wrr is publicly available for general usage.

\section*{Acknowledgments}
We'd like to thank Omer Levy for sharing his script for preprocessing the corpora used in the paper. We also thank the anonymous reviewers for their valuable comments and suggestions.

\bibliography{bibfile}

\begin{thebibliography}{}

\bibitem[\protect\citename{Agirre \bgroup et al.\egroup }2009]{AgiAlfHal09}
Eneko Agirre, Enrique Alfonseca, Keith Hall, Jana Kravalova, Marius Pasca, and
  Aitor Soroa.
\newblock 2009.
\newblock A study on similarity and relatedness using distributional and
  wordnet-based approaches.
\newblock {\em Proceedings of Human Language Technologies}, pages 19--27.

\bibitem[\protect\citename{Arora \bgroup et al.\egroup }2015]{AroLiLiaMaetal15}
Sanjeev Arora, Yuanzhi Li, Yingyu Liang, Tengyu Ma, and Andrej Risteski.
\newblock 2015.
\newblock Random walks on context spaces: Towards an explanation of the
  mysteries of semantic word embeddings.
\newblock Technical report, ArXiV.
\newblock \url{http://arxiv.org/pdf/1502.03520.pdf}.

\bibitem[\protect\citename{Bahdanau \bgroup et al.\egroup }2015]{BahChoBen15}
Dzmitry Bahdanau, Kyunghyun Cho, and Yoshua Bengio.
\newblock 2015.
\newblock Neural machine translation by jointly learning to align and
  translate.
\newblock In {\em Proceedings of the International Conference on Learning
  Representations (ICLR)}.

\bibitem[\protect\citename{Bartlett \bgroup et al.\egroup }2006]{JorBarMcA06}
Peter~L. Bartlett, Michael~I. Jordan, and Jon~D. McAuliffe.
\newblock 2006.
\newblock Convexity, classification, and risk bounds.
\newblock {\em Journal of the American Statistical Association},
  101(473):138--156.

\bibitem[\protect\citename{Bottou and Bousquet}2011]{BotBou11}
L{\'e}on Bottou and Olivier Bousquet.
\newblock 2011.
\newblock The tradeoffs of large-scale learning.
\newblock {\em Optimization for Machine Learning}, page 351.

\bibitem[\protect\citename{Bruni \bgroup et al.\egroup }2012]{BruBolBar12}
Elia Bruni, Gemma Boleda, Marco Baroni, and Nam~Khanh Tran.
\newblock 2012.
\newblock Distributional semantics in technicolor.
\newblock {\em Proceedings of the 50th Annual Meeting of the Association for
  Computational Linguistics}, pages 136--145.

\bibitem[\protect\citename{Collobert and Weston}2008]{ColWes08}
Ronan Collobert and Jason Weston.
\newblock 2008.
\newblock A unified architecture for natural language processing: Deep neural
  networks with multitask learning.
\newblock In {\em Proceedings of the 25th international conference on Machine
  learning}, pages 160--167. ACM.

\bibitem[\protect\citename{Cover and Thomas}1991]{CovTho91}
T.~M. Cover and J.~A. Thomas.
\newblock 1991.
\newblock {\em Elements of Information Theory}.
\newblock John Wiley and Sons, New York.

\bibitem[\protect\citename{Ding and Vishwanathan}2010]{DinVis10}
Nan Ding and S.~V.~N. Vishwanathan.
\newblock 2010.
\newblock $t$-logistic regression.
\newblock In Richard Zemel, John Shawe-Taylor, John Lafferty, Chris Williams,
  and Alan Culota, editors, {\em Advances in Neural Information Processing
  Systems 23}.

\bibitem[\protect\citename{Dongarra \bgroup et al.\egroup
  }1990]{DonCroDufHam90}
J.~J. Dongarra, J.~Du Croz, S.~Duff, and S.~Hammarling.
\newblock 1990.
\newblock A set of level 3 basic linear algebra subprograms.
\newblock {\em {ACM} Transactions on Mathematical Software}, 16:1--17.

\bibitem[\protect\citename{Finkelstein \bgroup et al.\egroup
  }2002]{FinGabMat02}
Lev Finkelstein, Evgeniy Gabrilovich, Yossi Matias, Ehud Rivlin, Zach Solan,
  Gadi Wolfman, and Eytan Ruppin.
\newblock 2002.
\newblock Placing search in context: The concept revisited.
\newblock {\em ACM Transactions on Information Systems}, 20:116--131.

\bibitem[\protect\citename{Gemulla \bgroup et al.\egroup }2011]{GemNijHaaSis11}
R.~Gemulla, E.~Nijkamp, P.~J. Haas, and Y.~Sismanis.
\newblock 2011.
\newblock Large-scale matrix factorization with distributed stochastic gradient
  descent.
\newblock In {\em Conference on Knowledge Discovery and Data Mining}, pages
  69--77.

\bibitem[\protect\citename{Hill \bgroup et al.\egroup }2014]{HilReiKor14}
Felix Hill, Roi Reichart, and Anna Korhonen.
\newblock 2014.
\newblock Simlex-999: Evaluating semantic models with (genuine) similarity
  estimation.
\newblock {\em Proceedings of the Seventeenth Conference on Computational
  Natural Language Learning}.

\bibitem[\protect\citename{Larochelle and Hinton}2010]{LarHin10}
Hugo Larochelle and Geoffrey~E. Hinton.
\newblock 2010.
\newblock Learning to combine foveal glimpses with a third-order boltzmann
  machine.
\newblock In {\em Advances in Neural Information Processing Systems (NIPS) 23},
  pages 1243--1251.

\bibitem[\protect\citename{Lee and Lin}2013]{LeeLin13}
Ching-Pei Lee and Chih-Jen Lin.
\newblock 2013.
\newblock Large-scale linear ranksvm.
\newblock {\em Neural Computation}.
\newblock To Appear.

\bibitem[\protect\citename{Levy and Goldberg}2014]{LevGol14}
Omer Levy and Yoav Goldberg.
\newblock 2014.
\newblock Neural word embedding as implicit matrix factorization.
\newblock In Max Welling, Zoubin Ghahramani, Corinna Cortes, Neil Lawrence, and
  Kilian Weinberger, editors, {\em Advances in Neural Information Processing
  Systems 27}, pages 2177--2185.

\bibitem[\protect\citename{Levy \bgroup et al.\egroup }2015]{LevGolDag15}
Omer Levy, Yoav Goldberg, and Ido Dagan.
\newblock 2015.
\newblock Improving distributional similarity with lessons learned from word
  embeddings.
\newblock {\em Transactions of the Association for Computational Linguistics},
  3:211--225.

\bibitem[\protect\citename{Liu \bgroup et al.\egroup }2015]{LiuJiaWei15}
Quan Liu, Hui Jiang, Si~Wei, Zhen{-}Hua Ling, and Yu~Hu.
\newblock 2015.
\newblock Learning semantic word embeddings based on ordinal knowledge
  constraints.
\newblock In {\em Proceedings of the Annual Meeting of the Association for
  Computational Linguistics (ACL)}, pages 1501--1511.

\bibitem[\protect\citename{Luong \bgroup et al.\egroup }2013]{LuoSocMan13}
Minh-Thang Luong, Richard Socher, and Christopher~D. Manning.
\newblock 2013.
\newblock Better word representations with recursive neural networks for
  morphology.
\newblock {\em Proceedings of the Seventeenth Conference on Computational
  Natural Language Learning}, pages 104--113.

\bibitem[\protect\citename{Maaten and Hinton}2008]{MaaHin08}
L.~van~der Maaten and G.E. Hinton.
\newblock 2008.
\newblock Visualizing high-dimensional data using t-sne.
\newblock {\em jmlr}, 9:2579--2605.

\bibitem[\protect\citename{Manning \bgroup et al.\egroup }2008]{ManRagSch08}
C.~D. Manning, P.~Raghavan, and H.~Sch{\"u}tze.
\newblock 2008.
\newblock {\em Introduction to Information Retrieval}.
\newblock Cambridge University Press.

\bibitem[\protect\citename{Mikolov \bgroup et al.\egroup
  }2013a]{MikCheCorDea13}
Tomas Mikolov, Kai Chen, Greg Corrado, and Jeffrey Dean.
\newblock 2013a.
\newblock Efficient estimation of word representations in vector space.
\newblock {\em arXiv preprint arXiv:1301.3781}.

\bibitem[\protect\citename{Mikolov \bgroup et al.\egroup
  }2013b]{MikSutCheCoretal13}
Tomas Mikolov, Ilya Sutskever, Kai Chen, Greg Corrado, and Jeffrey Dean.
\newblock 2013b.
\newblock Distributed representations of words and phrases and their
  compositionality.
\newblock In Chris Burges, Leon Bottou, Max Welling, Zoubin Ghahramani, and
  Kilian Weinberger, editors, {\em Advances in Neural Information Processing
  Systems 26}.

\bibitem[\protect\citename{Mnih \bgroup et al.\egroup }2014]{MniHeeGra14}
Volodymyr Mnih, Nicolas Heess, Alex Graves, and Koray Kavukcuoglu.
\newblock 2014.
\newblock Recurrent models of visual attention.
\newblock In {\em Advances in Neural Information Processing Systems (NIPS) 27},
  pages 2204--2212.

\bibitem[\protect\citename{Pennington \bgroup et al.\egroup }2014]{PenSocMan14}
Jeffrey Pennington, Richard Socher, and Christopher~D Manning.
\newblock 2014.
\newblock Glove: Global vectors for word representation.
\newblock {\em Proceedings of the Empiricial Methods in Natural Language
  Processing (EMNLP 2014)}, 12.

\bibitem[\protect\citename{Radinsky \bgroup et al.\egroup }2011]{RadAgiGab11}
Kira Radinsky, Eugene Agichtein, Evgeniy Gabrilovich, and Shaul Markovitch.
\newblock 2011.
\newblock A word at a time: Computing word relatedness using temporal semantic
  analysis.
\newblock {\em Proceedings of the 20th international conference on World Wide
  Web}, pages 337--346.

\bibitem[\protect\citename{Rockafellar}1970]{Rockafellar70}
R.~T. Rockafellar.
\newblock 1970.
\newblock {\em Convex Analysis}, volume~28 of {\em Princeton Mathematics
  Series}.
\newblock Princeton University Press, Princeton, NJ.

\bibitem[\protect\citename{Usunier \bgroup et al.\egroup }2009]{UsuBufGal09}
Nicolas Usunier, David Buffoni, and Patrick Gallinari.
\newblock 2009.
\newblock Ranking with ordered weighted pairwise classification.
\newblock In {\em Proceedings of the International Conference on Machine
  Learning}.

\bibitem[\protect\citename{Vilnis and McCallum}2015]{VilMcc15}
Luke Vilnis and Andrew McCallum.
\newblock 2015.
\newblock Word representations via gaussian embedding.
\newblock In {\em Proceedings of the International Conference on Learning
  Representations (ICLR)}.

\bibitem[\protect\citename{Wang \bgroup et al.\egroup }2015]{WanLinTsv15}
Ling Wang, Chu-Cheng Lin, Yulia Tsvetkov, Silvio Amir, Ramon~Fernandez
  Astudillo, Chris Dyer, Alan Black, and Isabel Trancoso.
\newblock 2015.
\newblock Not all contexts are created equal: Better word representations with
  variable attention.
\newblock In {\em EMNLP}.

\bibitem[\protect\citename{Weston \bgroup et al.\egroup }2012]{WesWanWeiBer12}
Jason Weston, Chong Wang, Ron Weiss, and Adam Berenzweig.
\newblock 2012.
\newblock Latent collaborative retrieval.
\newblock {\em arXiv preprint arXiv:1206.4603}.

\bibitem[\protect\citename{Yin and Kushner}2003]{YinKus03}
G~George Yin and Harold~Joseph Kushner.
\newblock 2003.
\newblock {\em Stochastic approximation and recursive algorithms and
  applications}.
\newblock Springer.

\bibitem[\protect\citename{Yun \bgroup et al.\egroup }2014]{YunRamVis14}
Hyokun Yun, Parameswaran Raman, and S.~V.~N. Vishwanathan.
\newblock 2014.
\newblock Ranking via robust binary classification and parallel parameter
  estimation in large-scale data.
\newblock In {\em nips}.

\end{thebibliography}
\bibliographystyle{emnlp2016}

\clearpage
\appendix

\twocolumn[
\begin{center}
\Large{\wrr: Learning Word Embeddings via Robust Ranking}\\
\vskip 0.05in
\large{\color{red}(Supplementary Material)}\\
\vskip 0.05in
\small{Shihao Ji, Hyokun Yun, Pinar Yanardag, Shin Matsushima, and S. V. N. Vishwanathan}
\vskip 0.3in
\end{center}]

\section{Parallel \wrr}
\label{sec:ParallelAlgorithm}

Given number of $p$ workers, we partition words $\Wcal$ into $p$ parts \{$\Wcal^{(1)}, \Wcal^{(2)}, \cdots, \Wcal^{(p)}\}$ such that they are
mutually exclusive, exhaustive and approximately equal-sized. This partition
on $\Wcal$ induces a partition on $\Ub$, $\Omega$ and $\Xi$ as
follows: $\Ub^{(q)} := \cbr{ \ub_w }_{w \in \Wcal^{(q)}}$,
$\Omega^{(q)} := \cbr{(w,c) \in \Omega}_{w \in \Wcal^{(q)}}$, and
$\Xi^{(q)} := \cbr{\xi_{(w,c)}}_{(w,c) \in \Omega^{(q)}}$ for $1
\leq q \leq p$. When the algorithm starts, $\Ub^{(q)}$,
$\Omega^{(q)}$ and $\Xi^{(q)}$ are distributed to worker $q$.

At the beginning of each outer iteration, an approximately
equal-sized partition $\{\Ccal^{(1)}$, $\Ccal^{(2)}, \cdots, \Ccal^{(p)}\}$
on the context set $\Ccal$ is sampled; note that this is
independent of the partition on words $\Wcal$. This induces a
partition on context vectors $\Vb^{(1)},\Vb^{(2)},\cdots,\Vb^{(p)}$
defined as follows: $\Vb^{(q)} := \cbr{ \vb_{c} }_{c \in \Ccal^{(q)}}$
for each $q$. Then, each $\Vb^{(q)}$ is distributed to each worker
$q$. Now we define
\begin{align}
  &\overline{J}^{(q)}(\Ub^{(q)}, \Vb^{(q)}, \Xi^{(q)}) =\nonumber\\
  \mkern-50mu &\sum_{(w,c)
    \in \Omega \cap \rbr{\Wcal^{(q)} \times \Ccal^{(q)}}} \sum_{c' \in
    \Ccal^{(q)} \setminus \cbr{c}}\mkern-18mu j(w,c,c'),
  \label{eq:part_obj}
\end{align}
where $j(w,c,c')$ was defined in \eqref{eq:def_jwc}. Note that
$j(w,c,c')$ in the above equation only accesses
$\ub_w$, $\vb_c$ and $\vb_{c'}$ which belong to no sets other than
$\Ub^{(q)}$ and $\Vb^{(q)}$, therefore worker $q$ can run stochastic
gradient descent updates on \eqref{eq:part_obj} for a predefined
amount of time without having to communicate with other workers. The
pseudo-code is illustrated in Algorithm~\ref{alg:parallel}.

Considering that the scope of each worker is always confined to a rather
narrow set of observations
$\Omega \cap \rbr{\Wcal^{(q)} \times \Ccal^{(q)}}$, it is somewhat
surprising that \newcite{GemNijHaaSis11} proved that such an optimization
scheme, which they call stratified stochastic gradient descent (SSGD),
converges to the same local optimum a vanilla SGD would converge to. This is due to the fact that
\begin{align}
  &\EE\Big[
    \overline{J}^{(1)}(\Ub^{(1)}, \Vb^{(1)}, \Xi^{(1)})
    +
    \overline{J}^{(2)}(\Ub^{(2)}, \Vb^{(2)}, \Xi^{(2)})
    + \nonumber\\
    &\cdots\mkern-3mu + \mkern-3mu
    \overline{J}^{(p)}(\Ub^{(p)}, \Vb^{(p)}, \Xi^{(p)})
  \Big]
  \approx
  \overline{J}(\Ub, \Vb, \Xi),
\end{align}
if the expectation is taken over the sampling of the partitions of
$\Ccal$. This implies that the bias in each iteration due to
narrowness of the scope will be washed out in a long run; this
observation leads to the proof of convergence in
\newcite{GemNijHaaSis11} using standard theoretical results from
\newcite{YinKus03}.

\begin{algorithm}
  \begin{algorithmic}
    \STATE {$\eta$: step size}
    \REPEAT 
    \STATE {\texttt{//Start outer iteration}}
    \STATE {Sample a partition over contexts $\Ccal^{(1)},
        \cdots\mkern-3mu, \Ccal^{(q)}$}
    \STATE {\texttt{//Step 1:Update $\Ub, \Vb$ in parall.}}
    
    \FORALLP {$q \in \cbr{1,\cdots,p}$}
    \STATE {Fetch all $\vb_c \in \Vb^{(q)}$}
    \REPEAT
    \STATE {Sample $(w,c)$ uniformly from $\Omega^{(q)} \cap
      \rbr{\Wcal^{(q)}\mkern-3mu \times\mkern-3mu \Ccal^{(q)}}$}
    \STATE {Sample $c'$ uniformly from $\Ccal^{(q)} \setminus \cbr{c}$}
    \STATE {\texttt{//following three updates are done simultaneously}}
    \STATE {$\ub_w\mkern-3mu \leftarrow\mkern-3mu \ub_w - \eta \cdot r_{w,c} \cdot \rho'(\xi^{-1}_{w,c})
      \cdot \ell'\rbr{\inner{\ub_{w}}{\vb_{c}\mkern-3mu -\mkern-3mu \vb_{c'}}} \cdot
      (\vb_c\mkern-3mu -\mkern-3mu \vb_{c'})$}
    \STATE {$\vb_c\mkern-3mu \leftarrow\mkern-3mu \vb_c - \eta \cdot r_{w,c} \cdot \rho'(\xi^{-1}_{w,c})
      \cdot \ell'\rbr{\inner{\ub_{w}}{\vb_{c}\mkern-3mu -\mkern-3mu \vb_{c'}}} \cdot
      \ub_w$} 
    \STATE {$\vb_{c'}\mkern-3mu \leftarrow\mkern-3mu \vb_{c'} + \eta \cdot r_{w,c} \cdot \rho'(\xi^{-1}_{w,c})
      \cdot \ell'\rbr{\inner{\ub_{w}}{\vb_{c}\mkern-3mu -\mkern-3mu \vb_{c'}}} \cdot
      \ub_w$}     
    \UNTIL {predefined time limit is exceeded}
    \ENDFOR
    \STATE {\texttt{//Step 2: Update $\Xi$ in parallel}}

    \FORALLP {$q \in \cbr{1,\cdots,p}$}
    \STATE {Fetch all $\vb_c \in \Vb$}
    \FOR {$w \in \Wcal^{(q)}$}
    \FOR {$c \in \Ccal$}
    \STATE {$\mkern-20mu\xi_{w,c}\mkern-3mu =\mkern-3mu \alpha / \rbr{\sum_{c'\in \Ccal \setminus \cbr{c}}
      \ell\rbr{\inner{\ub_{w}}{\vb_{c}\mkern-3mu -\mkern-3mu \vb_{c'}}}\mkern-3mu +\mkern-3mu \beta}$}
    \ENDFOR
    \ENDFOR
    \ENDFOR
    \UNTIL {$\Ub$, $\Vb$ and $\Xi$ are converged}
  \end{algorithmic}
  \caption{Distributed \wrr algorithm.}
  \label{alg:parallel}
\end{algorithm}%\vspace{-0.3cm}

\section{Additional Experimental Details}
\label{sec:AdditExperDeta}

Table \ref{tab:scale_exp} is the tabular view of the data plotted in Figure \ref{fig:scale_exp} to provide additional experimental details. 

\begin{table*}[htb]
\caption {Performance of \wv, \gl and \wrr on datasets with increasing
  sizes; evaluated on WS-353 word similarity  benchmark and Google word
  analogy benchmark.} 
\label{tab:scale_exp}
\begin{center}
\begin{threeparttable}
\begin{tabular}{|l||c|c|c||c|c|c|}
\hline          Corpus Size & \multicolumn{3}{c||}{WS-353 (Word Similarity)} & \multicolumn{3}{c|}{Google (Word Analogy)} \\
\cline{2-7}     & \wv & \gl & \wrr & \wv & \gl & \wrr\\
  \hline 17M  & 66.8 & 47.8 & 70.4 & 39.2 & 30.4 & 44.5 \\
  \hline 32M  & 64.1 & 47.8 & 68.4 & 42.3 & 30.9 & 52.1 \\
  \hline 64M  & 67.5 & 55.0 & 70.8 & 53.5 & 42.0 & 59.9 \\
  \hline 128M & 70.7 & 54.5 & 72.8 & 59.8 & 50.4 & 65.1 \\
  \hline 256M & 72.0 & 59.5 & 72.4 & 67.6 & 60.3 & 68.6 \\
  \hline 512M & 72.3 & 64.5 & 74.1 & 70.6 & 66.4 & 70.6 \\
  \hline 1.0B & 73.3 & 68.3 & 74.0 & 70.4 & 68.7 & 70.8 \\
  \hline 1.6B & 71.8 & 69.5 & 74.1 & 72.1 & 70.4 & 71.7 \\
  \hline 7.2B & 73.4 & 70.9 & 75.2/77.4$^1$ & 75.1$^2$ & 75.6$^2$ & 76.0$^{2,3}$ \\
  \hline
\end{tabular}%\vspace{-0.3cm}
\begin{tablenotes}
  \scriptsize\item $^1$ When $\rho_0$ is used, corresponding to setting $\xi\mkern-3mu=\mkern-3mu 1$ in training and no $\xi$ update
  \scriptsize\item $^2$ Use 3CosMul instead of regular 3CosAdd for evaluation
  \scriptsize\item $^3$ Use $\ub_w$ instead of default $\ub_w+\vb_c$ as word representation for evaluation
\end{tablenotes}
\end{threeparttable}
\end{center}
\end{table*}

\end{document}